\documentclass[final]{article}
\usepackage{tabularx}
\usepackage{arxiv}
\usepackage[utf8]{inputenc} % allow utf-8 input
\usepackage[T1]{fontenc}    % use 8-bit T1 fonts
\usepackage{hyperref}       % hyperlinks
\usepackage{url}            % simple URL typesetting
\usepackage{booktabs}       % professional-quality tables
\usepackage{amsfonts}       % blackboard math symbols
\usepackage{nicefrac}       % compact symbols for 1/2, etc.
\usepackage{microtype}      % microtypography
\usepackage{lipsum}		% Can be removed after putting your text content
\usepackage{graphicx}
\usepackage{lineno}
\usepackage{mathtools}
\usepackage[most]{tcolorbox}
\usepackage{authblk}
\usepackage[title]{appendix}
\usepackage{url}
\usepackage{setspace}
\usepackage[superscript,biblabel]{cite}

\newcommand{\beginsupplement}{%
        \setcounter{table}{0}
        \renewcommand{\thetable}{S\arabic{table}}%
        \setcounter{figure}{0}
        \renewcommand{\thefigure}{S\arabic{figure}}%
     }

\title{Decontextualized learning for interpretable hierarchical representations of visual patterns}

\author[1, 2, 3, \thanks{\texttt{Corresponding author: ietheredge@ab.mpg.de}}]{
  R. Ian Etheredge}
\author[4, 5, 6, 7]{
  Manfred Schartl}
 \author[1, 2, 3]{Alex Jordan}

\affil[1]{Department of Collective Behaviour, Max Planck Institute of Animal Behavior, Konstanz, Germany}

\affil[2]{Centre for the Advanced Study of Collective Behaviour, University of Konstanz, Konstanz, Germany}

\affil[3]{Department of Biology, University of Konstanz, Konstanz, Germany}

\affil[4]{Centro de Investigaciones Científicas de las Huastecas Aguazarca, A.C., Calnali, Hidalgo, Mexico}

\affil[5]{Developmental Biochemistry, Biocenter, University of Würzburg, Würzburg, Bavaria, Germany}

\affil[6]{Hagler Institute for Advanced Study, Texas A\&M University, College Station, TX, USA}

\affil[7]{Xiphophorus Genetic Stock Center, Texas State University San Marcos, San Marcos, TX, USA}

\begin{document}
% \linenumbers

\maketitle

\begin{abstract}
Apart from discriminative models for classification and object detection tasks, the application of deep convolutional neural networks to basic research utilizing natural imaging data has been somewhat limited; particularly in cases where a set of interpretable features for downstream analysis is needed, a key requirement for many scientific investigations. We present an algorithm and training paradigm designed specifically to address this: decontextualized hierarchical representation learning (DHRL). By combining a generative model chaining procedure with a ladder network architecture and latent space regularization for inference, DHRL address the limitations of small datasets and encourages a disentangled set of hierarchically organized features. In addition to providing a tractable path for analyzing complex hierarchal patterns using variation inference, this approach is generative and can be directly combined with empirical and theoretical approaches. To highlight the extensibility and usefulness of DHRL, we demonstrate this method in application to a question from evolutionary biology.


\end{abstract}

\section{Introduction}

The application of deep convolutional neural networks (CNNs \cite{LeCun2010}) to supervised tasks is quickly becoming ubiquitous, even outside of standardized visual classification tasks.\cite{Everingham2010} In the life sciences, researchers are leveraging these powerful models for a broad range of domain-specific discriminative tasks such as automated tracking of animal movement \cite{Stephens2017,Mathis2018,Pereira2019,Graving2019}, the detection and classification of cell lines~\cite{falk2019u,riba2020single,mcquin2018cellprofiler}, and mining genomics data~\cite{poplin2018universal}.

A key motivation for the expanded use of deep feed-forward networks lies in their capacity to capture increasingly abstract and robust representations. However, outside of the objective function they have been optimized on, building interpretability into these representations is often difficult as networks naturally absorb all correlations found in the sample data and the features which are useful for defining class boundaries can become highly complex (Figure \ref{fig:convolution}). For many investigations the main objective falls outside of a clearly defined detection or classification task, e.g. identifying a set of descriptive features for downstream analysis, and interpretability and generalizability is much more important. Because of this, in contrast to many traditional computer vision algorithms \cite{hough1962method, harris1988combined, Lowe1999, Lowe2004}, the application of more expressive approaches built on CNNs and other deep networks to research has been limited \cite{Ezray2019} (Figure \ref{fig:compare}).

Unsupervised learning, a family of algorithms designed to uncover unknown patterns in data without the use of labeled samples, offers an alternative for compression, clustering, and feature extraction using deep networks. Generative modeling techniques have been especially effective in capturing the complexity of natural images, i.e. generative adversarial networks (GANs \cite{Goodfellow2014}) and variational autoencoders (VAEs, \cite{Kingma2014, Rezende2014}). VAEs in particular offer an intuitive way for analyzing data. As an extension of variational inference, VAEs combine an inference model, which performs amortized inference (typically a CNN) to approximate the true posterior distribution and encode samples into a set of latent variables ($q_{\phi}(z|x)$), and a generative model which generates new samples from those latent variables ($p_{\theta}(x|z)$). Instead of optimizing on a discriminative task, the objective function in VAEs is less strictly defined but typically seeks to minimize the reconstruction error between inputs $x$ and outputs $p_{\theta}(q_{\phi}(x))$ (reconstruction loss) as well as the divergence between the distribution of latent variables $q_{\phi}(z|x)$ and the prior distribution $p(z)$ (latent regularization). 

\subsection{Overcoming Hurdles to Application}
\begin{table}
   \caption{Desired characteristics of an integrative tool for investigations of natural image data and general representation learning meta-prior enforcement strategies.}\label{tab:metapriors}
\begin{tabularx}{\linewidth}{>{\centering\let\newline\\\arraybackslash\hspace{0pt}}X>{\centering\let\newline\\\arraybackslash\hspace{0pt}}X>{\centering\let\newline\\\arraybackslash\hspace{0pt}}X}

\toprule
Desired Characteristic & Representation Learning Meta-Prior\cite{Bengio2013} & Example Approach\\\\

\midrule
Disentangling factors of variation & Limited number of shared factors of variation  & Latent regularization\cite{Zhao2017, Gretton2008}\\\\
Capturing spatial relationships & Hierarchical organization of representation   & Hierarchical model architecture\cite{Zhao2018}\\\\
Incorporating existing knowledge & Local variation on manifolds & Structured latent codes\cite{Chen2016}\\\\
Connect analyses and experiments & Local variation on manifolds & Generative models\cite{Kingma2014,Goodfellow2014}\\\\
Inference &  Probability mass and local variation on manifolds & Variational inference\cite{Kingma2014}

\end{tabularx}
\end{table}

In VAEs, two problems often arise which are of primary concern to researchers using natural imaging data. \textit{1)} The mutual information between $x$ and $z$ can become vanishingly small, resulting in an uninformative latent code and overfit to sample data, the information preference problem;\cite{Chen2018,Zhao2018} this  is particularly true when using powerful convolutional decoders which are needed to create realistic model output.\cite{Chen2016, Chen2018, Zhao2017} \textit{2)} In contrast to the hierarchical representations produced by deep feed-forward networks used for discriminative tasks, in generative models local feature contexts become emphasized at the cost of large-scale spatial relationships. This is a product of the restrictive mean-field assumption of pixel-wise comparisons and produces generative models capable of reproducing complex image features while using only local feature contexts without capturing higher-order spatial relationships within the latent encoding.\cite{Zhao2018} 

The basis of a more expressive and robust approach for investigating natural image data has some key requirements: \emph{1)} provide a useful representation which disentangles factors of variation along a set of interpretable axes; \emph{2)} capture feature contexts and hierarchical relationships; \emph{3)} incorporate existing knowledge of feature importance and relationships between samples; \emph{4)} allow for statistical inference of complex traits; and \emph{5)} provide direct connections between analytical, virtual and experimental approaches. Here we integrate meta-prior enforcement strategies taken from represnetation learning \cite{Bengio2013} to specifically address the requirements of researchers using natural image data (Table \ref{tab:metapriors}).

Here we propose to address the limitations of existing approaches and incorporate the specific requirements of researchers using a combination of meta-prior enforcement strategies. VAEs with a ladder network architecture has been show to better capture a hierarchy of feature by mitigating the explain away problem of lower level feature, allowing for bottom-up and top-down feedback.\cite{Zhao2018} Additionally, combining pixel-wise error with a perceptual loss function \cite{Johnson2016} adapted from neural style transfer \cite{Gatys2015, Gatys2015b}, may also reduce the restrictive assumptions of amortized inference and pixel-wise reconstruction error by balancing them against abstract measures of visual similarity. 

\begin{figure}
   \includegraphics[width=\textwidth]{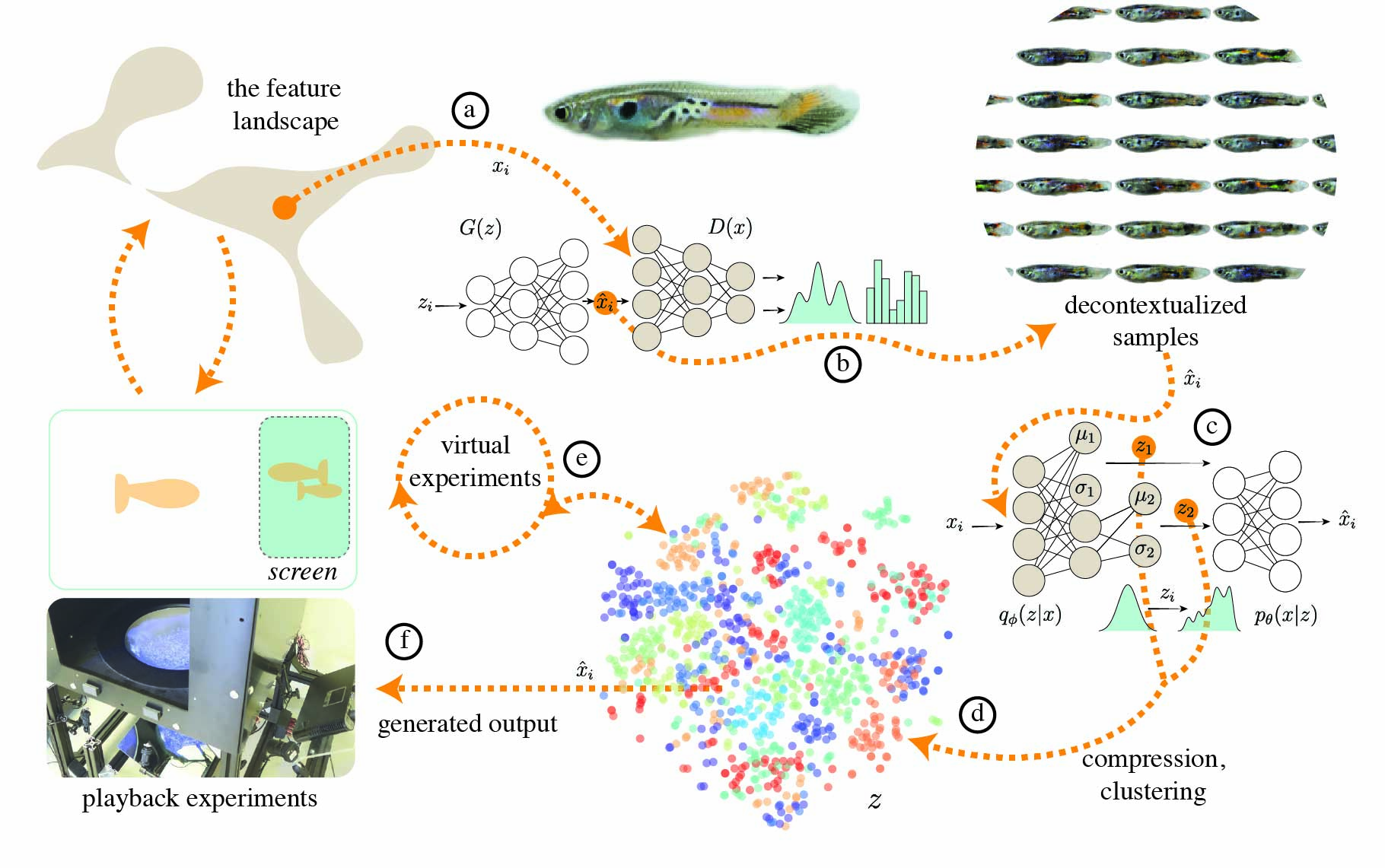}
   \caption{\emph{Overview}. a) Many patterns (e.g. male guppy ornaments) consist of combinations of several elements which have hierarchical relationships, spatial dependence, and feature contexts which may hold distinct biological importance. In our proposed framework, small sample sizes are supplemented using a generative (GAN) model which learns images statistics sufficient to produce novel out of sample examples (b). This model can be used to produce an unlimited number of novel samples and also reduces the covariances within sample data, which is advantageous for disentangling generative features. We use these "decontextualizd" generated outputs as input (c) to a variational auto encoder (VAE). Via a specific combination of meta-prior enforcement strategies and network architectures, we capture the hierarchical structure, disentangling factors of variation across multiple scales of increasing abstraction ($z_1$ through $z_n$). Using the learned distribution over these variables, the latent representation,  parameterized by a mean and variance term, we (d) define  a color-pattern space. Using this low dimensional representation we can (e) interface with downstream models such as evolutionary algorithms and (f) produce photo-realistic outputs to be used in playback experiments and immersive VR. Interpolations through the color-pattern space with animated models and VR allows researchers to manipulate generated output for experimental tests. Techniques represented with dashed lines: (d) capturing a hierarchy of visual features, (e) combining a low-dimensional latent representation with virtual experiments and (f) playback experiments represent the current gaps in our analytical and experimental framework. Our approach directly address these shortcomings to span these gaps, creating a robust, integrated framework for investigating natural visual stimuli.}
   \label{fig:methods}
\end{figure}

In terms of the latent regularization, a disentangled representation of causal factors requires an information-preserving latent code. Choosing a regularization techniques which mitigate the trade off between inference and data fit\cite{Gretton2008} can encourage the disentanglement of generative factors along a set of variables in an interpretable way. We also propose a novel training paradigm inspired by GAN chaining that further relaxes the natural covariances in the data: \textit{decontextualized learning} and actually uses the restrictive assumptions of GAN generator networks to our advantage to overcome the limitations of small datasets, typical for many studies in the natural sciences and further increase the disentanglement of generative factors (Figure \ref{fig:methods}, Methods \ref{methods:keymethods}). 

While several metrics have been proposed for assessing interpretability and disentanglement \cite{Higgins2017, Kim2018, Eastwood2018}, these metrics rely heavily on the associated labels, well defined features or stipulations from classification of detection competitions, e.g. \cite{dauphin2012unsupervised}. In addition to being highly domain specific, for most practical investigations in the natural sciences, these types of labels do not exist and we must often rely on fundamentally qualitative assessments. In many cases, labeled data is not available and interpreting traversals of the latent code (Figure \ref{fig:vlae_trav}) may introduce our own perceptual biases. Here, we adapt an approach from explainable AI: integrated gradients \cite{sundararajan2017axiomatic} in application to latent variable exploration too provide a direct assessment of latent variables, quantifying latent feature attributions without the necessity of labeled data and allows for exploring latent variables without adding additional human biases (Methods \ref{methods:lfa}).

We demonstrate the proposed framework using two example datasets: male guppy ornamentation and butterfly wing patterns from the discipline of sensory ecology and evolution (see Appendix \ref{evolutionbackground} for motivation and background on existing approaches).

\section{Results}

\begin{figure}

   \includegraphics[width=\textwidth]{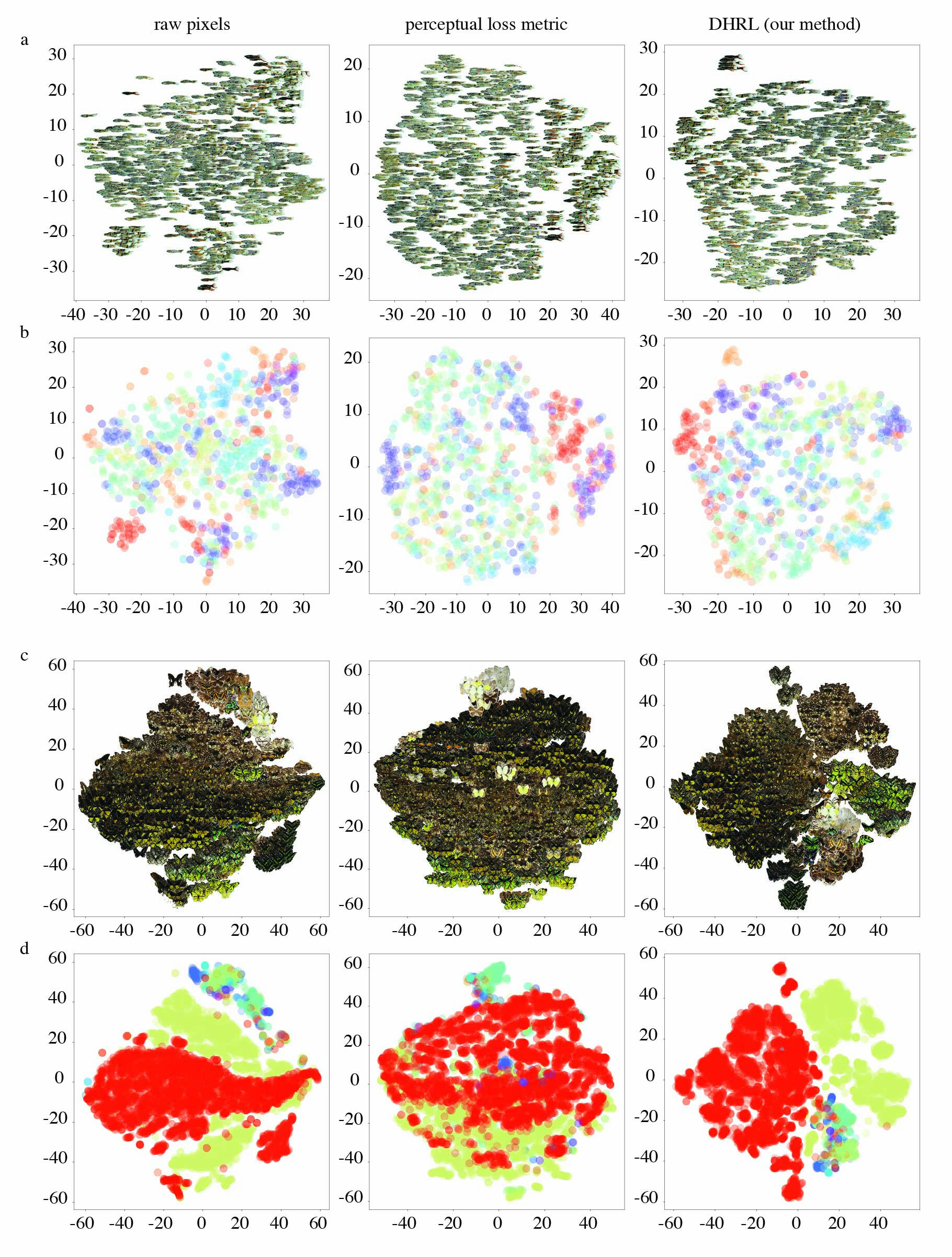}
   \caption{\emph{Comparing with existing techniques}. 2-dimensional embedding of (left) raw pixel distributions, (middle) using a perceptual similarity score (\cite{Johnson2016, Ezray2019}), and (right) our framework. a) guppies, b) butterflies. Colors indicate unique sub groups for each sample (guppy variety and butterfly species).} 
   \label{fig:compare}
\end{figure} 

While biological datasets are typically small, they are usually highly structured and standardized compared to large classification datasets (e.g. ImageNet\cite{ImageNet}). This provides an advantage for controlling noise and uninformative covariates in the data. Using a modified infoGAN\cite{Chen2016} architecture, we incorporate prior knowledge about the structure of our sample data to generate realistic samples from the complex image distribution conditioned on a set of latent variables. Here, we incorporate prior knowledge about our samples of male guppy ornamentation images by providing a 32-class categorical latent code (Figure 3b, top right). These 32-classes represent the 32 individual tanks, unique subsets of the overall sample, with shared traits related to guppy ornamentation patterns. The categories learned by the trained model posses unique features which also covary in the sample data, e.g. a distinct black bar and orange stripe which characterizes one guppy species, \textit{P. wengei} (Figure \ref{fig:vlae_trav}, a). While generated samples share characteristics and even resemble known varieties, generated samples posses decontextualized combinations of features across examples (Figure \ref{fig:vlae_trav}, a). We use these, decontextualized samples as input to our variational (VAE) model for our "decontextualized" training paradigm.

\begin{figure}
  \includegraphics[width=\textwidth]{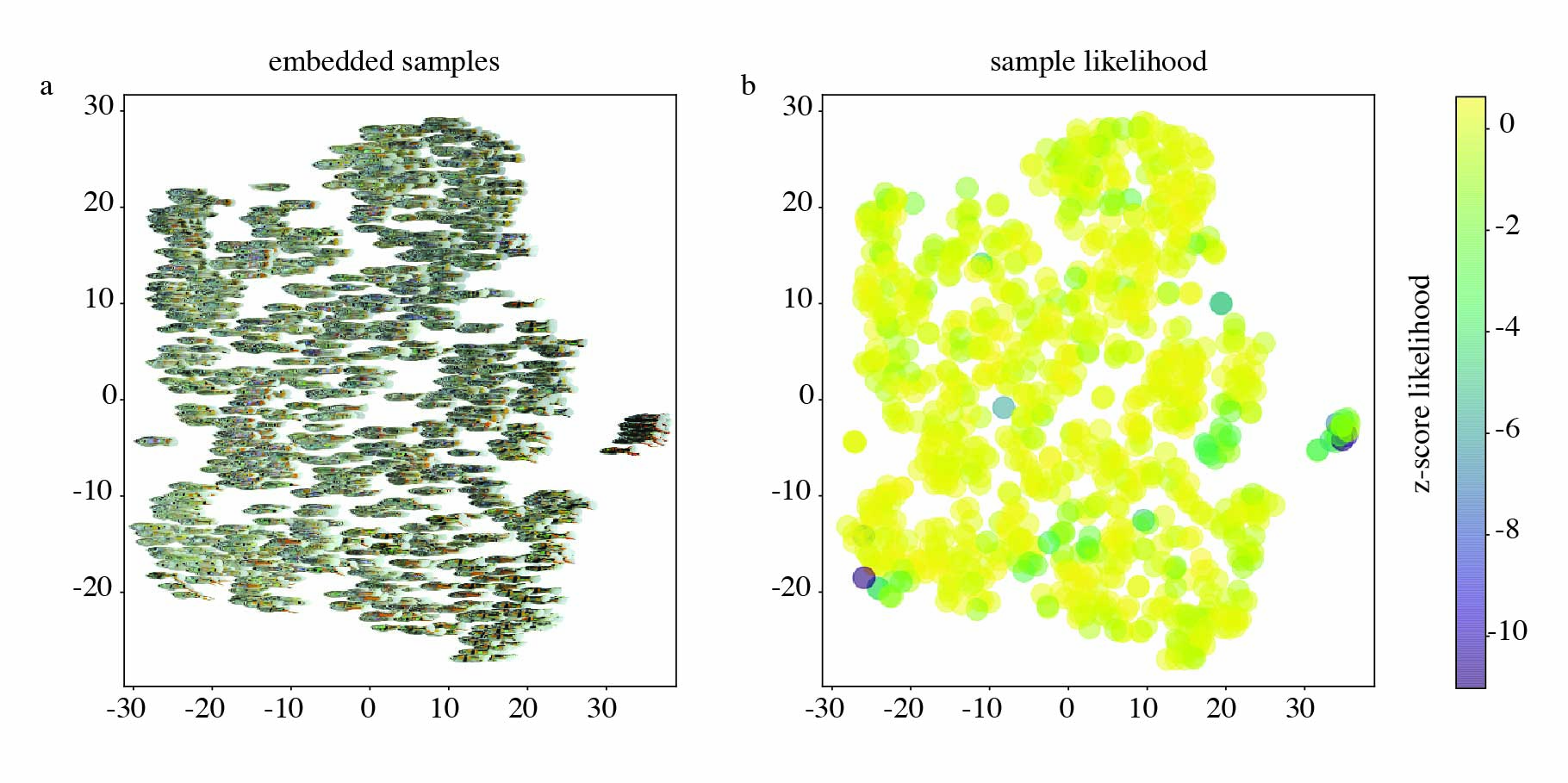}
  \caption{\emph{Sample likelihood estimates}. a) Embedded samples b) Normalized (standard score) likelihood estimates for each sample.} 
  \label{fig:vlae_lh}
\end{figure}

GAN training and VAE training are performed in separate steps so that models are not jointly optimized. The generated samples from the trained GAN model are used as training data to a variational model (Figure \ref{fig:methods}) with a hierarchical model architecture\cite{Zhao2018} which consists of 10 latent variables across four codes ($z_1, \dots, z_4$) with increasing expressivity, (Methods \ref{methods:vlae}). We observe distinct clusters in the latent space of the trained model which correspond to sample categories and differs qualitatively from two existing method (raw pixel and perceptuall loss embeddings using tSNE \cite{vanderMaaten2008, Kobak2019}, Figure \ref{fig:compare}). The unique latent space of the four latent encodings capture unique factors of variation in the sample data in a scale-dependent way (Figure \ref{fig:vlae_trav}, Figure \ref{fig:vlae_topo}). In this model, $z_1$, the latent code with the lowest capacity captures local traits such as the color and intensity of discrete patches, e.g. $z_{1_1}$ encodes variation in the intensity of an orange spot (\ref{fig:vlae_trav} 4b, left). At higher levels ($z_2, \dots, z_4$), latent variables encode complex traits which combine multiple elements, (\ref{fig:vlae_trav} 4b, right). We use this same latent representation to describe the relationship between samples and calculate likelihood estimates. Samples with rare traits, e.g. such as the “Tr5” strain in our sample data which are distinctly melanated, cluster together in the embedded space, and have a low sample likelihood (\ref{fig:vlae_lh}).

Embedding the 4, 10-dimensional, latent codes reveals scale-dependent relationships between elements. In $z_1$ (Figure \ref{fig:vlae_topo}, left) color values and local features dominate the relationship between points (Figure \ref{fig:vlae_topo}, left). Nearest neighbor samples (Minkowski distance \cite{Dubes1980} in the 10-dimensional space, Figure \ref{fig:vlae_topo}, b) show color similarity whereas higher order features, e.g. patterning and morphology, determine the relationships between samples in the more expressive latent spaces ($z_2,\dots,z_4$). Though we find strong covariance between features across scales, in some cases the nearest neighbors samples differ greatly depending on the scale and feature context (Figure \ref{fig:vlae_topo}, b). 

\begin{figure}
   \includegraphics[width=\textwidth]{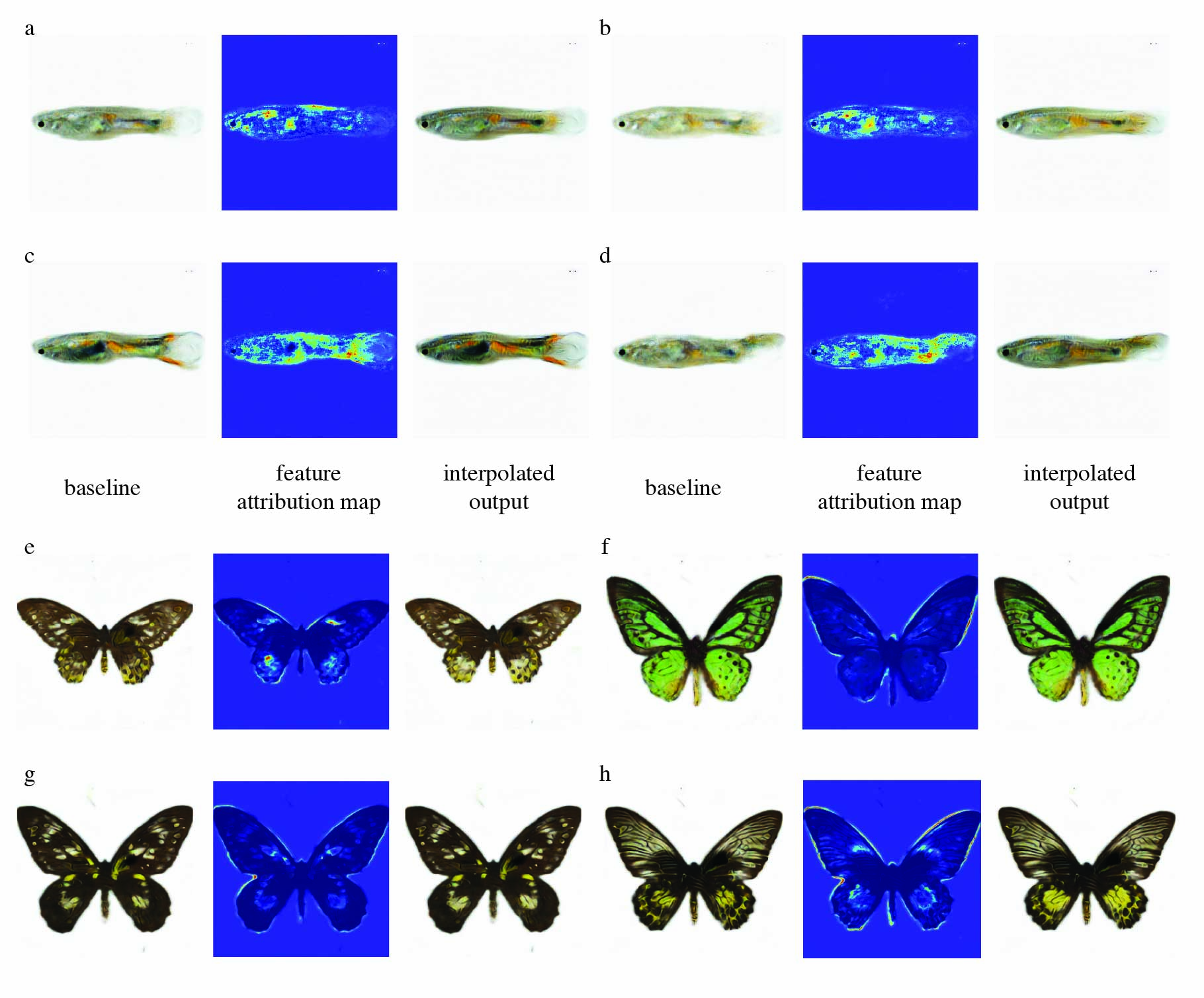}
   \caption{\emph{Latent variable feature attribution}. a-d) Four samples of genereated guppy images performing integrated gradeints feature attribution (see Methods \ref{methods:lfa}). a-d) guppy images visualizing the feature attributions of the latent variable $z_13$ of the DHRL trained variational model. e-h) butterfly images using latent feature $z_37$. Heatmap values have been normalized using a standard score. Images to the left are generated with the latent feature set to its lowest value in the sample, to the right with the highest value in the sample.}
   \label{fig:vlae_feature attribution}
\end{figure}

\begin{table}
   \caption{Disentanglement and completeness metrics for of VLAE inference network across datasets and when using our decontextualized learning approach (DHRL).}\label{tab:disentanglement}
\begin{tabularx}{\linewidth}{XXXXX}
\toprule
VLAE Training Data & $D(z_1)$, $C(z_1)$ & $D(z_2)$, $C(z_2)$ & $D(z_3)$, $C(z_3)$ & $D(z_4)$, $C(z_4)$\\\\
\midrule
Butterflies (n=9531) & 0.64, 0.60 & 0.67, 0.55 & 0.63, 0.51 & 0.88, 0.60\\\\
Guppies (n=987) & 0.29, 0.32 & 0.12, 0.13 & 0.13, 0.16 & 0.56, 0.66 \\\\
Guppies (gen., n=19k) & 0.12, 0.16 & 0.13, 0.18 & 0.32, 0.42 & 0.62, 0.75\\\\
Guppies \textbf{DHRL} & 0.14, 0.16 & 0.18, 0.23 &  0.31, 0.39 & \textbf{0.90, 0.95}
\end{tabularx}
\end{table}

We assess the level of disentanglement of our trained variational models using the metric established in \cite{Eastwood2018} using known class labels as attribute classes (butterfly species, learned class from infoGAN pre-training, and guppy strain varieties). Across models, we find the most expressive latent codes ($z_4$) provide the highest degree of disentanglement between known classes with the highest disentanglement score overall using our decontextualized, DHRL method (see Table \ref{tab:disentanglement}). 

We also provide a qualitative approach for attributing latent variables to image features using network gradients (Methods \ref{methods:lfa}); when labels are unknown. In Figure \ref{fig:vlae_feature attribution}, a-d we visualize one variable of $z_1$, the least expressive latent variable space ($z_{13}$) of the DHRL-trained guppy latent variable model. We find that the same latent variable controls the relative intensity of green color patches across individuals. Looking at a single variable of more expressive latent codes $z_{27}$ of the trained butterfly model (Figure \ref{fig:vlae_feature attribution}, e-h) we find that this latent variable controls the size of yellow patches on the lower wings relative to the size of yellow patches on the upper wings (when patches are not present this variable has no effect (Figure \ref{fig:vlae_feature attribution}, f). Further investigation of latent variables can be performed using the provided tool (\url{https://github.com/ietheredge/VisionEngine/notebooks/IntegratedGradients.ipynb}).

\begin{figure}
  \includegraphics[width=\textwidth]{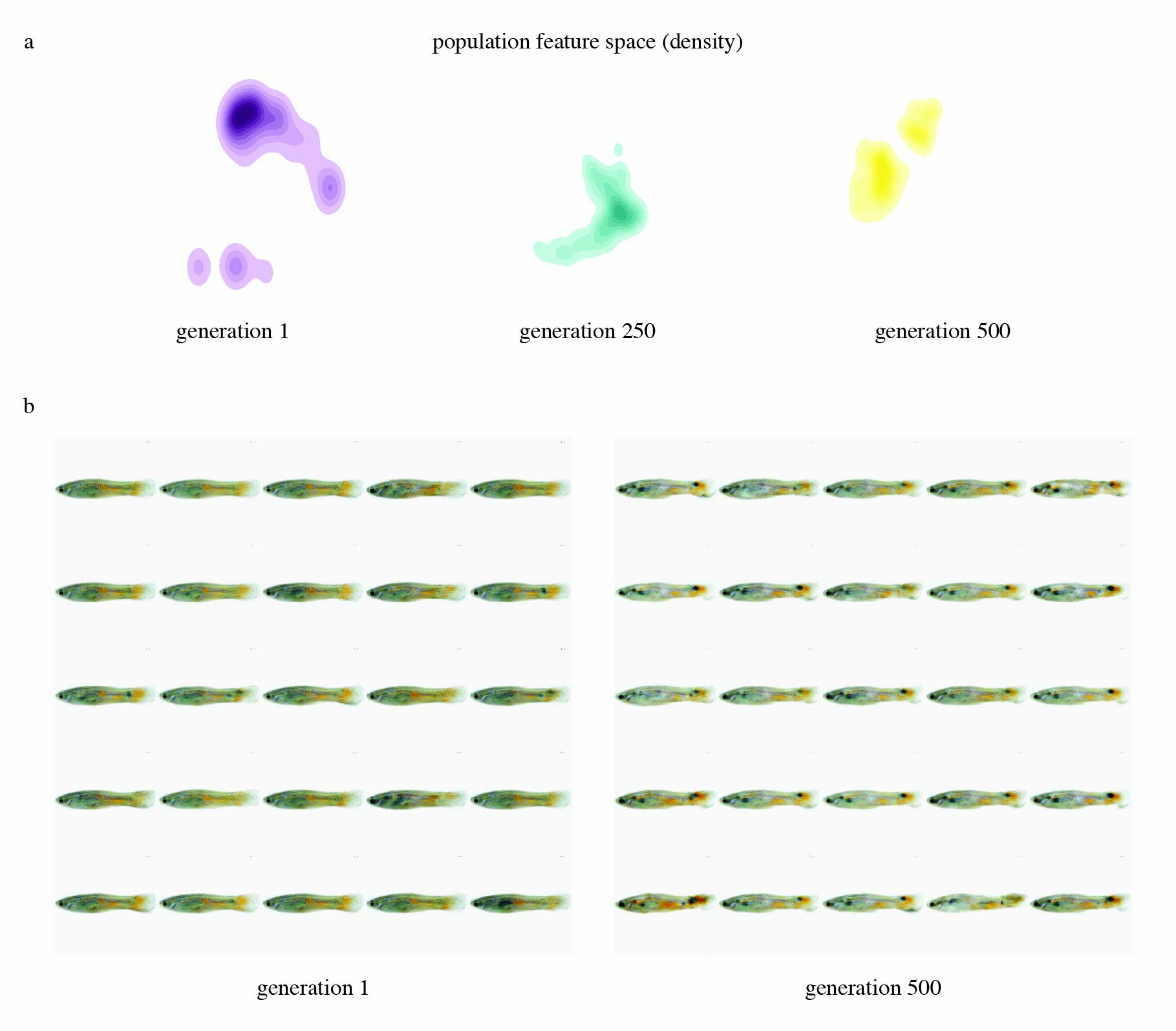}
  \caption{\emph{Virtual Experiments}. a) Kernel density plot of samples over generations 1, 250, 500 selecting orange ornaments and contrast. After 500 generations the population has shifted from the initial sample distribution, finding two peaks which maximize the fitness function. b) Samples of initial parent population, left, with the highest fitness, compared to those with the highest fitness after 500 generations. Samples in later generations show higher numbers of brighter orange and dark melanated patches and increased within body contrast.} 
  \label{fig:evolution}
\end{figure}

Using the latent representation, $z$, of our DHRL trained variational model of guppy ornaments as input, we apply an evolutionary algorithm (Figure \ref{fig:evolution}), defined by a fitness function from the guppy literature: oranger, higher contrast males are preferred by females.\cite{Houde1987} Starting from a parent population initialized by our sample embedding (900 samples), we simulate 500 generations under these selective forces. We observe exaggerated and more numerous orange and black patches in novel configurations compared to the initial population (Figure \ref{fig:evolution}, b). Projecting the latent representation of generations 1, 250, and 500, we find that instead of a single peak, after several generations, many novel solutions are optimized (Figure \ref{fig:evolution} a). Investigating the values of the latent variables over generations reveals two distinct latent factors driven to fixation in the population under these selective forces (\ref{fig:alleles}). We also observe to population optimization of latent factors over time in Movie \ref{fig:evospacemovie}. Using a single Titan Xp GPU with 12GB memory we could simulate a population size of 1000 individuals in an average of 19.5 seconds per generation. 

\section{Discussion}

Supervised discriminative learning algorithms are already becoming an integral tool for researchers across disciplines whereas unsupervised generative modeling approaches remain a relatively young and active area of machine learning research. Already, the highly expressive generative models like the ones presented here are transforming the way we interact with image data. By solving problems in a more general way, generative modeling approaches provide more direct connections to hypothesis testing and connecting observations. Here, we demonstrate how these approaches may serve as an engine for more integrative studies of animal coloration patterns, and natural image data more generally, directly connecting approaches. 

Analytically, our approach captures important hierarchical features across spatial scales that existing approaches do not account for (Figure \ref{fig:compare}, Figure \ref{fig:vlae_topo},  Appendix \ref{significance}), it removes the inherent biases of predefined filters by learning features directly from the sample data, and it disentangles complex factors of variation into a useful, meaningful representation (Figure \ref{fig:vlae_trav}, Figure \ref{fig:vlae_feature attribution}). More than compressing data into a low dimensional space, this approach is generative and can create novel out-of-sample examples with high fidelity. This is a potentially transformative extension for researchers in the natural sciences which is not offered by existing approaches, allowing researchers to test analytical results with virtual experiments, and empirically, by using virtual reality playback experiments or observational studies (see Movie \ref{fig:vrmovie}). 

These techniques can be adapted to many domain specific questions (see \ref{significance} for a specific discussion regarding the potential impact of this approach on the study of color pattern evolution). As the latency between input and output decreases in video playback experiments, integrating instantaneous behavioral feedback and in-the-loop methods for hypothesis testing may be used to design complex real-time assays. More sophisticated virtual experiments may also incorporate agent based models and evolutionary algorithms working directly on the latent representation to create complex simulations (e.g as in \cite{Ha2018}, Figure \ref{fig:evolution}). In our demonstration, we are able to simulate 1000 individuals in under 20 seconds per generation with very little optimization and asynchronous approaches may already be possible. Analytically, as research in machine learning aimed at understanding how information is organized and used by algorithms advances, a growing theoretical framework with a basis in statistical mechanics\cite{Bahri2020} and information theory\cite{Tishby2000} may provide additional avenues for investigating the statistical properties of color pattern spaces and their evolution.

\section{Experimental Procedures}\label{methods}

\subsection{Materials Availability}

Guppy images were collected from a maintained stock at the University of Wuerzburg under authorization 568/300-1870/13 of the Veterinary Office of the District Government of Lower Franconia, Germany, in accordance with the German Animal Protection Law (TierSchG). Individuals were imaged on a white background with fixed lighting conditions \cite{Kemp2008} using a Cannon D600 digital camera. Images were down sampled and center cropped to final size of 256 x 256 pixels. The dataset consists of 977 standardized RGB images across three species and 13 individual strains.

Butterfly images were downloaded from the Natural History Museum, London under a creative commons license (DOIs: https://doi.org/10.5519/qd.gvq3p7xq, https://doi.org/10.5519/qd.pw8srv43). This dataset consists of 9531 RGB images.

For each dataset, we segmented samples from the background using a customized object segmentation network adapted from \cite{Caelles2017}. For each dataset we annotated 8 samples to train the segmentation network. All samples were cropped and resized to 256 x 256 and placed on a transparent background (RGBA). For calculating the perceptual loss during training, images were translated to 3-channel images with a white background using alpha blending.

Updated links to original data repositories can be accessed here: \url{https://github.com/ietheredge/VisionEngine/README.md}.

\subsubsection{Data and Code Availability}

All models were implemented using Tensorflow 2.2 and can be accessed here: \url{https://github.com/ietheredge/VisionEngine}, including installation and evaluation scripts to reproduce our results. Instructions for creating new data loaders for training new datasets using this method can be found at \url{https://github.com/ietheredge/VisionEngine/data_loaders/datasets/README.md}.

\begin{figure}
   \includegraphics[width=\textwidth]{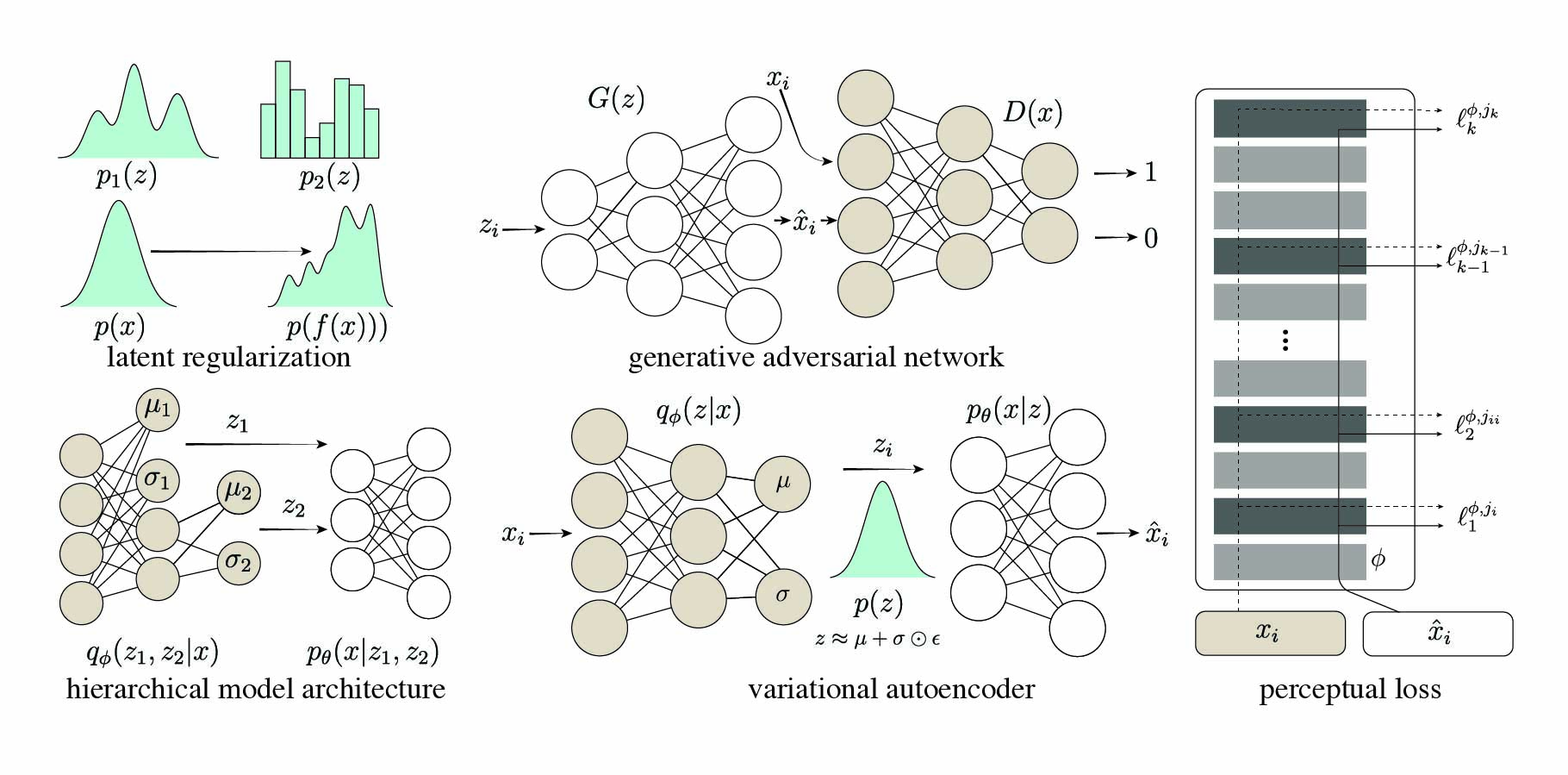}
   \caption{\emph{Key Methods} \emph{Top left}: the distributions of our latent representation may be parameterized by a number of continuous or discrete variables. In infoGAN, a categorical latent code is combined with continuous latent codes which allows for disentangling substructures in the sample data without labeled samples. \emph{Top right}, example structure of a generative adversarial network. Here, a noise vector, $z_i$ is input to the generator network $G(z)$ which produces a reconstructed output $\hat{x_i}$. A real sample, $x_i$, and generated sample $\hat{x_i}$ are subsequently passed through a separate discriminator network $D(x)$ which determines if the sample is real ($1$) or generated ($0$). In infoGAN, the latent encoding of generated samples is optimized an additional network $Q$ which shares all convolutional layers with $D$. \emph{Bottom left}, the generic architecture of a variational ladder autoencoder (VLAE). Multiple latent spaces $(z_1, z_2, ..., z_k)$ are learned with each successive input layer having increasing expressivity and abstraction (abilitiy to combine features across spatial scales). \emph{Bottom middle},  structure of a variational autoencoder (VAE). $x_i$ and $\hat{x_i}$ are an example input and its reconstructed output,  the probabilistic encoder or inference model, $q_\phi(z|x)$, performs posterior inference learning shared model parameters, $\phi$, across samples, approximating the true posterior distribution. The probabilistic decode, $p_\theta(Z|X)$, $p_\theta(X|Z)$, learns a joint distribution of the encoded space, $Z$, and the data space $x$. The low dimensional bottleneck, $Z$ is a distribution of latent variables capable of reconstructing sample inputs, parameterized by a vector of means $\mu$ and standard deviations $\sigma$. The noise term $\epsilon$ allows for the parameters of this multivariate distribution to be optimized using back propagation, known as the reparametrization trick. \emph{Right}, Perceptual loss models use a pretrained network, $\phi$, e.g. VGG-16\cite{Simonyan2014}. Two samples, the original input and reconstructed output, are input to the model and the maxpooling layer activations for each are used as outputs. The distance between these functions emphasizes higher-level similarity than standard pixel-wise differences. Outputs  from the shallow layer $\ell_1$ represents low-level local features where as output from the deeper layer $\ell_k$ contains information from across spatial scales and more abstract representations. The euclidean distance between these activation outputs gives a metric for the similarity of the two inputs as "perceived" by a network pre-trained on a much broader dataset. Perceptual loss functions can be used as a stand-alone transfer-learning approach to finding perceptual differences between samples or as part of any network as an additional or alternative reconstruction loss (see \ref{fig:compare}).}
   \label{fig:methods_details}
\end{figure}

\subsection{Key Methods}\label{methods:keymethods}

DHRL relies on a three-step process of sequential training where first a generative adversarial network is trained to transform a noise sample into realistic out of sample examples. Next, a variational autoencoder is pre-trained on the generated samples. Then finally, the pretrained variational model is fine-tuned on the original samples. 

\subsubsection{InfoGAN}\label{methods:infogan}

We use an unsupervised approach to disentangle discrete and continous latent factors adapted from \cite{Chen2016} (InfoGAN) which modifies the minimax game typically used for training GANs such that:

\begin{equation}
    \min _{G, Q} \max _{D} V_{I}(D, G, Q)=V(D, G)-\lambda L_I(G,Q)
\end{equation}

where $V(D, G)$ is the original GAN objective introduced in \cite{Goodfellow2014} and $L_I(G,Q)$ approximates the lower bound of the mutual information $I(c ; G(z, c))$ using Monte Carlo sampling such that $L_I(G,Q) \leq I(c ; G(z, c))$.\cite{Chen2016} Like the generator $G$ and discriminator $D$, $Q$ is parameterized as a neural network and shares all convolutional layers with $D$.  

Both discrete $Q(c_d|x)$ and continuous latent codes $Q(c_c|x)$ are provided with continuous latent codes treated as a factored Gaussian distributions. Importantly, InfoGAN does not require supervision and no labels are provided, e.g. \cite{Kim2018}. 

We substitute the original generator and discriminator models from \cite{Chen2016} with the architecture described in \cite{Redmon2016} and increase the flexibility of the latent code, providing additional continuous and discrete latent codes. For guppy experiments, we provide two continuous and 19 discrete codes (samples were drawn from 19 paternal lines). For the basis noise vector input to the generator, we used 100-unit random noise vector. 

\subsubsection{Variational Ladder Autoencoder}\label{methods:vlae}

In contrast to hierarchical architectures, e.g. \cite{bachman2016architecture, sonderby2016ladder}, we learn a hierarchy of features by using multiple latent codes with increasing levels of abstraction from \cite{Zhao2018}, i.e. $q_{\phi}(z_1, \dots, z_L|x)$ . The expressivity of $z_i$ is determined by its depth. The encoder  $q_{\phi}(z_1, \dots, z_L|x)$ consists of four blocks such that:

\begin{equation}
    H_{\ell}=G_{\ell}(H_{\ell-1})
\end{equation}
\begin{equation}
    z_{\ell}\sim\mathcal{N}(\mu_{\ell}(H_{\ell}), \mathbf{I})
\end{equation}

where $H_{\ell}$, $G_{\ell}$, and $\mu_{\ell}$ are neural networks. For our encoder model, $G_{\ell}$ is a stack of convolutional, batch normalization, and leaky rectified linear unit activation (Conv-BN-LeakyReLU), we stack four Conv-BN-LeakyReLU blocks  for each $G_{\ell}$ with increasing number of channels for each subsequent convolutional layer, i.e. N-channels/2, N-channels, N-channels, N-channels*2 where N-channels is 16, 64, 256, 1024 for $G_{1}, G_{2}, G_{3}, G_{4}$ respectively. We apply spectral normalization to all convolutional layers (see below). Because we want to preserve feature localization, we use average pooling followed by a squeeze-excite block to apply a context-aware weighting to each channel (see below). 

Similarly, the decoder, $p_{\theta}(x|z_1, \dots, z_L)$, is composed of blocks such that:

\begin{equation}
    \tilde{z}_{\ell}=U_{\ell}([\tilde{z}_{\ell+1};V_{\ell}(z_{\ell})])
\end{equation}

where $[.;.]$ denotes channel-wise concatenation. Parallel to $G_{\ell}$, blocks in the encoder: $U_{\ell}$ are composed of  Conv-BN-ReLU blocks (note the use of ReLU and not LeakyReLU in the decoder) with decreasing number of channels in each convolutional layer, i.e. N-channels*2, N-channels, N-channels, N-channels/2 where N-channels is 1024, 256, 64, 16. No spectral normalization wrappers or squeeze-excite layers are applied in the decoder.

\subsubsection{Squeeze-Excite Layers}\label{methods:selayers}

Squeeze-and-Excitation Networks \cite{hu2018squeeze} were proposed to improve feature interdependence by adaptively weighting each channel within a feature map based on the filter relevance by applying a a channel-wise recalibaration. Here we apply squeeze-excite (SE) layers prior to the variational layer such that each embedding $z_i$ captures features with cross-channel dependencies. Each SE layer consists of a global average pooling layer which averages channel-wise features followed by two fully connected layers with relu activations, the first with size channels/16 and the second with the same size as the number of input channels. Finally a sigmoid, "excite," layer assigns channel wise probabilities which are then multiplied channel wise with the original inputs.

\subsubsection{Reconstruction Loss}\label{methods:reconstructionloss}

We minimize the negative log likelihood of the sample data by minimizing the mean squared error between input and output, jointly optimizing the reconstruction loss for each sample $x$:

\begin{equation}
\begin{split}
\mathcal{L}_{\text {pixel-wise}}=\mathbb{E}_{p_{\text {data}}}(x) \mathbb{E}_{q_{\phi}}(z \mid x)\left[\log p_{\theta}(x \mid z)\right]
\\
=\frac{1}{n} \sum_{i=1}^{n}\left(x_{i}-p_{\theta}(q_{\phi}(x_{i}\right)))^{2}
\end{split}
\end{equation}

To relax the restrictive mean-field assumption which is implicit in minimizing the pixel-wise error, we jointly optimize the similarity between inputs and outputs using intermediate layers of a pretrained network, VGG16 \cite{Simonyan2014}, as feature maps.\cite{Gatys2015,Gatys2015b,Johnson2016} Here we calculate the Gram matrices of feature maps, which match the feature distributions of real and generated outputs for each layer as:

\begin{equation}
G_{a b}^{\ell}=\frac{\sum_{c d} F_{c d a}^{\ell}(x) F_{c d b}^{\ell}(x)}{C D}
\end{equation}

\begin{equation}
\mathcal{L}_{\text {perceptual}}=\sum_{\ell=1}^{L}\frac{\frac{1}{n} \sum_{i=1}^{n}\left(G_{a b}^{\ell}(x_i) - G_{c d}^{\ell}(p_{\theta}(q_{\phi}(x_{i})))\right)^{2}}{L}
\end{equation}

for feature maps $F_{a}$ and $F_{b}$ in layer $\ell$ across locations $c$, $d$. This measures the correlation between image filters and is equivalent to minimizing the distance between the distribution of features across feature maps, independently of feature position.\cite{li2017demystifying}

The combined reconstruction loss is a weighted sum of the perceptual loss and pixel-wise error:

\begin{equation}
\mathcal{L}_{\text {reconstruction}}=\alpha\mathcal{L}_{\text {perceptual}}+\beta\mathcal{L}_{\text {pixel-wise}}
\end{equation}

where $\alpha$ and $\beta$ are Lagrange multipliers controlling the influence of each loss term. Here we set $\alpha$ = 1e-6 and $\beta$ = 1e5 to balance the contribution of reconstruction terms with variational loss (see below).

\subsubsection{Maximum Mean Discrepancy}\label{methods:mmd}

We use the maximum mean discrepancy approach (MMD)\cite{Gretton2008} to maximize the similarity between the statistical moments of $p(z)$ and $q_\phi(x)$ using the kernel embedding trick:

\begin{equation}
\operatorname{MMD}(p(z) \| q_{\phi}(z))=\mathbb{E}_{p(z), p\left(z^{\prime}\right)}\left[k\left(z, z^{\prime}\right)\right]+\mathbb{E}_{q_{\phi}(z), q_{\phi}\left(z^{\prime}\right)}\left[k\left(z, z^{\prime}\right)\right]-2 \mathbb{E}_{p(z), q_{\phi}\left(z^{\prime}\right)}\left[k\left(z, z^{\prime}\right)\right]
\end{equation} 

using a Gaussian kernel, $k\left(z, z^{\prime}\right)$, such that

\begin{equation}
k\left(z, z^{\prime}\right)=e^{-\frac{\left\|z-z^{\prime}\right\|^{2}}{2 \sigma^{2}}}
\end{equation}

to measure the similarity between $p_\theta(z)$ and $q_\phi(z)$ in Euclidean space. We measured similarity using multiple kernels with varying degrees of smoothness, controlled by the value of $\sigma^2$, i.e. multi-kernel MMD, \cite{gretton2012optimal} with varying bandwidths: $\sigma^2$ = 1e-6, 1e-5, 1e-4, 1e-3, 1e-2, 1e-1, 1, 5, 10, 15, 20, 25, 30, 35, 100, 1e3, 1e4, 1e5, 1e6.

Weighing the influence of MMD kernel differences on the combined objective function is controlled by the Lagrange multiplier $\lambda$ applied across each latent code. Giving the combined objective:

\begin{equation}
\mathcal{L}_{\text {total}}=\left(\sum_{i}^{L}\lambda\operatorname{MK-MMD}\left(q_\phi(z_i) \| p(z_i)\right)\right)+\mathcal{L}_{\text {reconstruction}}
\end{equation}

where $L$ is the number of hierarchical latent codes and $z_i$ is the n-dimensional latent code and the prior, $p(z_i) = \mathcal{N}(0, \mathbf{I})$ and $\mathcal{L}_{\text {reconstruction}}$ define above. Here, we set $\lambda$ = 1.

\subsubsection{Denoise Training}\label{methods:denoising}

In addition to further relaxing the contribution of pixel-wise error, adding a denoising criterion has been shown to yield better sample likelihood by learning to map both training data and corrupted inputs to the true posterior, providing more robust training for out of sample data.\cite{Im2017} We implement this with the addition of noise layer which samples a corrupted input $\tilde{x}$ from input $x$ before passing $\tilde{x}$ to the encoder $q_{\phi}(z|\tilde{x})$. We use apply random binomial noise (salt and pepper) to ten percent of pixels.

\subsubsection{Spectral Normalization}

Spectral normalization has been proposed as a method to prevent exploding gradients when using rectified linear units to stabilize GAN training via a global regularization on the weight matrix of each layer as opposed to gradient clipping to provide bounded first derivatives (the Lipschitz constraint) \cite{miyato2018spectral}. 

\subsection{Latent Feature Attribution and Disentanglement}\label{methods:lfa}

Understanding the importance of features for model predictions is an active area of research. Integrated gradients, introduced by \cite{sundararajan2017axiomatic},  assigns feature importance, determining causal relationships between predictions and image features by summing the gradients along paths between $x^{\prime}$ and $x$. 

\begin{equation}
\text {IG}_{i}(x)::=\left(x_{i}-x_{i}^{\prime}\right) \times \int_{\alpha=0}^{1} \frac{\partial \mathcal{P}\left(x^{\prime}+\alpha \times\left(x-x^{\prime}\right)\right)}{\partial x_{i}} d \alpha
\end{equation}

We adapt this procedure to investigate the contribution of each latent variable parameter $z_i$ where we use a baseline $z$, an encoding of a singe sample $x$ and iterate $z_j$ while holding all other $z_l$ constant and summing the gradients of the decoder $p_{\theta}(x|z)$ such that:

\begin{equation}
\text {IG}_{i}^{approx}(p_{\theta}(x|z^j))::=\left(p_{\theta}(x|z^j)_i-p_{\theta}(x|z^{{j}^{\prime}})_i\right) \times \sum_{k=1}^{m} \frac{\partial \mathcal{P}\left(p_{\theta}(x|z^{{j}^{\prime}}): z^{{j}^{\prime}}_j = z^{{j}^{\prime}}_j +\frac{k}{m} \times (z^{j}_j-z^{{j}^{\prime}}_j)\right)}{\partial p_{\theta}(x|z^j)_{i}} \times \frac{1}{m}
\end{equation}

where $j$ is the axis of latent code being interpolated, $i$ is the individual feature (pixel), $p_{\theta}(x|z)$  is the reconstructed output, $p_{\theta}(x|z^{\prime})$ is the baseline reconstructed output, $k$ is the perturbation constant, and $m$ is the number of steps in the approximation of the integral. We use the Riemann sum approximation of the integral over the interpolated path $\mathcal{P}$ which involves computing the gradient in a loop over the inputs for $k=1,\dots,m$. Here, we use $m$ = 300 and $k = 2\max(|z|)$ for each $z^{j}$ starting from a baseline $p_{\theta}(x|z^{{j}^{\prime}}) : z_j = -\max(|z|)$.

We use the technique developed in \cite{Eastwood2018} for assessing disentanglement, measuring the relative entropy of latent factors for predicting class labels. We measure disentanglement of $D_i$ of each latent code is measured by $D_i = (1-H_K(P_i))$ where $H_K$ is the entropy and $P_i$ is the relative importance of the generative factor. We also include a metric of completeness $C_i$, approximating the degree to which the generative factor is captured by a single latent variable, where $C_j = (1-H_D(P_j))$ where $P_j$ is the unweighted contribution of generative factors.\cite{Eastwood2018} Here, in the absence of labeled features, we use species (butterflies), breeding line variants (guppies), and predicted class of the generative model (generated guppies, \ref{methods:infogan}, above) for each model as approximate class labels (one class). This approximation naturally overestimates $D_i$ and underestimates $C_j$ as there is some overlap between classes in terms of visual features (see Figure \ref{fig:compare}, Figure \ref{fig:vlae_topo}). While \cite{Eastwood2018} proposes a third term to evaluate representations $I$ to measure the relative informativeness, we found that this value was highly coupled to the choice of the Lagrange multiplier $\lambda$ used for latent regularization (above).

\subsection{Simulating Evolution on the Latent Space}\label{methods:evolution}

For demonstrating an example virtual experiment, we use a genetic algorithm, with a parent population of 1000 random samples, evolved over 500 generations. Parent samples are random initialized across the the latent variables of each latent code. Fitness was calculate as an equally weighted sum of the total percentage of pixels within two ranges (orange rgb(0.9, 0.55, 0.) > rgb(1., 0.75, 0.1) and black rgb(0., 0., 0.) < rgb(0.2, 0.2, 0.2)) measured on the generated output, a simplification of empirical results from the literature.\cite{Houde1987,HoudeEndler1995} During each generation predicted fitness for each sample in the population was measured by the fitness of the nearest neighboring value in the reference table (for processing speed). To simulate weak selective pressure on the fitness function, we drew 500 random parent subsamples weighted by their proportional fitness. An additional 200 samples were drawn, without the proportional fitness weighting. Together, from the 700 subsamples in each generation we drew 300 random pairs, the "alleles" from each sample (the specific latent variable values) were chosen randomly with equal probability to create a combined offspring between the two samples. Each combined offspring then had two alleles randomly mutated, one by drawing from a random normal distribution and the other by replacing an existing value with zero (similar to destabilizing and stabilizing mutations). The next generation thus consisted of 100 samples, 700 parent samples + 300 offspring. This process was repeated for 500 generations.

\section{Acknowledgments}\label{acknowledge}
We would like to thank members of the Dept. of Collective Behavior, Max Planck Institute of Animal Behavior and  Centre for the Advanced Study of Collective Behaviour, University of Konstanz for comments on earlier versions of the manuscript as well as the Max Planck Computing and Data Facility for use of computational resources.

\subsection{Author contributions}
 RIE conceived the approach and designed the methodology; MS and RIE collected sample data. RIE wrote the manuscript. AJ secured funding. All authors contributed to editing and approving the manuscript. 
 
\subsection{Declaration of Interests}
 The Authors have no financial or non-financial competing interest.

\bibliographystyle{unsrt}
\bibliography{references}

\begin{appendices}

\beginsupplement

\section{Example Application to the Evolution of Color Patterns: Background}\label{evolutionbackground}

The incredible variety of color patterns seen in nature evolved under the selective forces imposed by the environment, and the visual experience of their receivers.\cite{Bates1863, Darwin1871, Wallace1877, Muller1878, Poulton1890, Thayer1909, Cott1940} Quantifying this diversity, and reliably testing the functional significance of these traits is fundamental to understanding fitness landscapes\cite{Smith1970} and underlies many subdisciplines of sensory ecology, cognitive neuroscience, collective behavior, and evolution. 

Creating quantitative descriptions of color patterns which take into account the unique sensory and semiotic worlds of their receivers\cite{von1992} has been a central challenge in visual ecology. Many tools have been developed: Quantitative Colour Pattern Analysis\cite{vandenBerg2019}, PAVO,\cite{Maia2019} Natural Pattern Match,\cite{Stoddard2014} among others.\cite{Gawryszewski2018, Tedore2016, Endler1991, Endler2005, Endler2012, Troscianko2015, Caves2017, Endler2018, Stoddard2019} Each of these tools uses one or an ensemble of complimentary metrics from image analysis and computer vision, e.g. image statistics, edge detection, and landmark-based filters.\cite{Lowe2004}

\begin{figure}[ht]
\centering
   \includegraphics[width=\textwidth/2]{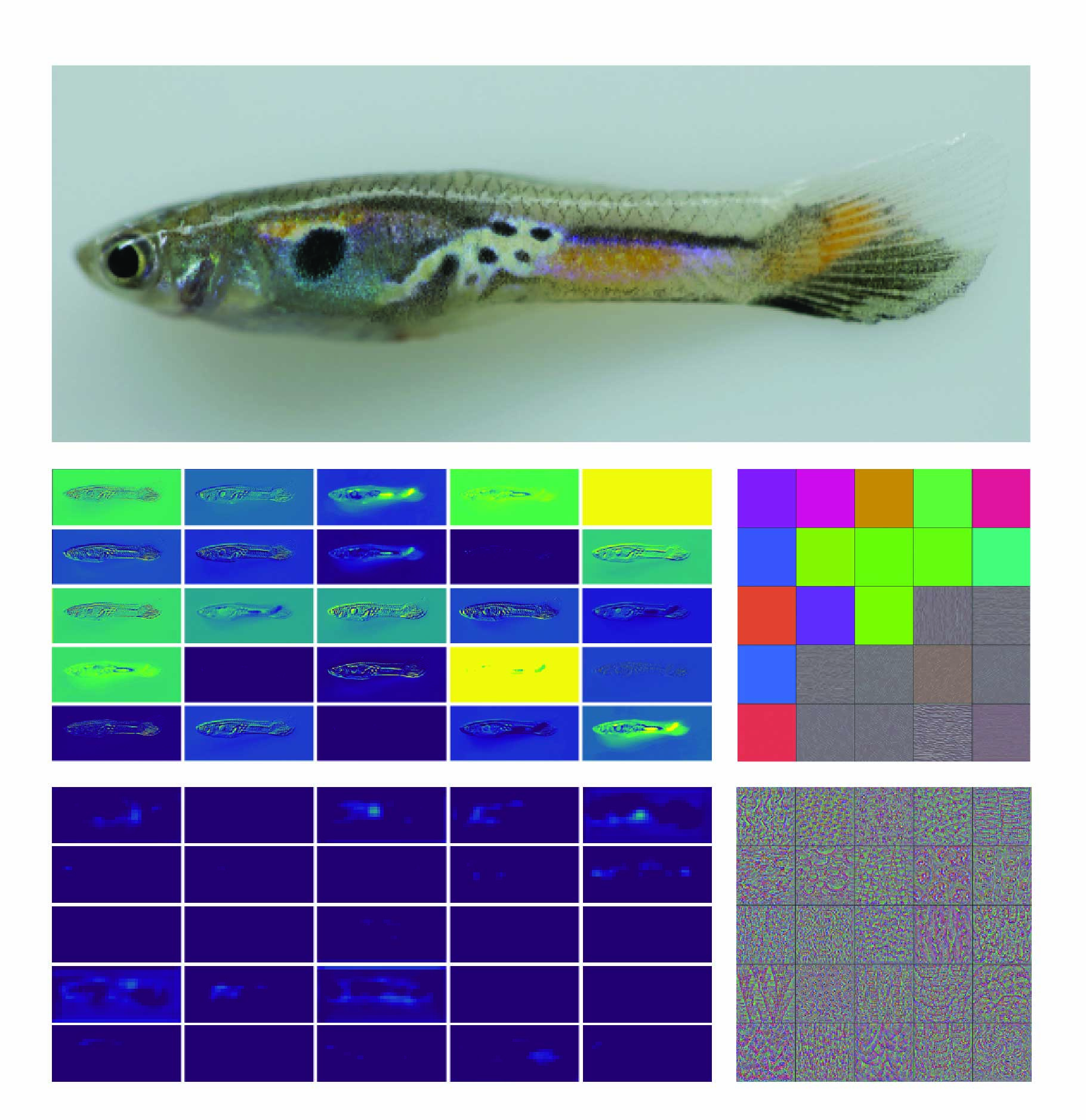}
   \caption{\emph{Convolutional layers.} In typical \emph{supervised} discriminative models, the objective being optimized is well defined, e.g. accurate classification or localization. As such, the representations provided by the downstream convolutional layers of deep networks take on characteristics optimized for task performance. At higher and higher network layers, the boundaries between classes can become complex and specialized to this objective because of the usefulness of such representations to identifying complex boundaries.Middle: Features learned at lower layers relate to color patches or gestures whereas at higher levels (bottom) fears become complex and interpretability can be difficult. Left, image pixels which are activated by pretrained image filters (yellow represents higher activations). Right, the maximally activating image feature for each filter. } 
   \label{fig:convolution}
  \end{figure}

Still, fundamental gaps remain. One of these gaps is the difficulty in building quantitative descriptions of complex features with multiple subelements. Most existing approaches fail to capture the full complexity of many of color patterns; the algorithms themselves are insufficiently expressive. This is particularly true when spatial or scale dependent relationships between features exist, e.g. the irregular patterns of male guppy ornamentation or butterfly wing patterns where similar sets of elements are arranged in species-specific configurations.\cite{Nijhout2001} Recently, researchers have begun employing machine learning algorithms such a as non-linear dimensionality reduction, e.g. t-distribute stochastic neighbor embedding (t-SNE,\cite{vanderMaaten2008, Kobak2019} Figure \ref{fig:compare}), and deep neural networks (Figure \ref{fig:compare},\cite{Ezray2019} Figure \ref{fig:convolution}). Still, while these techniques can better represent more complex relationships between pixel values within an image, current implementations do not disentangle features across scales or provide extensions to downstream experiments.

While complex trait may be difficult to quantify, they are nonetheless biologically relevant in terms of feature context\cite{Fechner1840,Fuller2002,Cole2016,Nieder2002, Fujita2012, Yang2016} and the perceptions of shape, motion, and attention\cite{Thayer1909, Cott1940, Kelley2014, Merilaita2017, Gasparini2013, Poulton1890}. And in the brain, we know that perception is hierarchically organized,\cite{Marr1982} and representations made at higher levels of the visual cortex and its homologs heavily influence the perception of low-level features.\cite{HubelWiesel1962, Pafundo2016} While measuring local features across an image provides important insight on regularity and the nature of wide-field variation, a collection of local feature descriptions across space is fundamentally different to a feature description built across scales.

Another gap is in building direct connections between approaches. Establishing spectral sensitivity, acuity, and feature importance is typically done using stimulus playback experiments or behavioral assays. However, beyond using statistical descriptions of features to guide researchers in the creation of stimuli there are few explicit connections between analysis and experiment. The current state of the art: immersive virtual reality (VR) and low-latency playback experiment---with fully animated, photo-realistic, 3D models, provide a rich experimental basis for investigating the relationship between visual inputs, neural activity, and behavior.\cite{Ingly2015, Stowers2017, Naik2020, Huang2020} VR systems are also beginning to better account for species-specific sensory biases including photoreceptor sensitivity, flicker fusion rate, acuity, and depth perception.\cite{Stowers2014, Naik2020} Still, currently these approaches rely on human-in-the-loop interventions for creating stimulus with even moderate complexity. 

Additionally, because color pattern traits have evolved under selective pressure from multiple receivers, establishing these types of evolutionary trade-offs is important to our understanding. However, experimental approaches often require large, highly disruptive manipulations such as translocation experiments or large scale crossbreeding experiments. Simulations and virtual experiments may better allows researchers to be explicit about the stimulus that is being tested and greatly reduce the number of subjects needed (Methods \ref{methods:evolution}). 

\subsection{The potential impacts of this approach on the study of evolution}\label{significance}

 This platform may be used to address many outstanding questions regarding the functional significance of color pattern traits; here, we discuss some of these questions. \emph{1)} What are the constraints on the evolvability of a given trait? By identifying the topographical relationship between different traits within the color pattern space we can test predictions about the selective forces acting on them related to their geometric relationships, e.g. the axes of variation in traits meant to communicate viability should show increased orthogonality compared to co-occurring traits which have evolved under a Fisherian process.\cite{darwin1859,fisher1915,lande1981,kirkpatrick1982,Iwasa1994,prum2010} \emph{2)} Categorical perception is an important perceptual mechanism for understanding the evolution of color signals.\cite{Caves2018} But in systems where color patterns are used for mimicry\cite{Bates1863, Wallace1877, Joron1998} or novelty, investigating the boundaries between complex traits is fundamental. By performing traversals across the distribution of the latent variables, interpolating between samples can allow for tests of continuous~\cite{Searcy2005} versus categorical perception\cite{Roff2015} of complex traits. \emph{3)} Many color pattern traits have evolved under selective pressure from multiple receivers, e.g. both females and predators shape the diversity of male guppy ornaments.\cite{Endler1987} Establishing these types of evolutionary trade-offs is difficult and often requires large, highly disruptive manipulations such as translocation experiments. Using evolutionary models similar to the ones presented here researchers can simulate multiple fitness landscapes and evolutionary trajectories simultaneously to perform a broad range of virtual experiments. Importantly, while each of these examples place either analytical, experimental, or virtual results at the center, by using the platform presented here, they maintain direct connections across approaches. Furthermore, they can incorporate existing techniques \cite{Endler1991, Endler2005, Endler2012, Troscianko2015, Caves2017, Endler2018, Stoddard2019} as image preprocessing routines, during playback, or constraints on virtual experiments.

\section{Supplemental Figures}

   \begin{figure}[ht]
   \includegraphics[width=\textwidth]{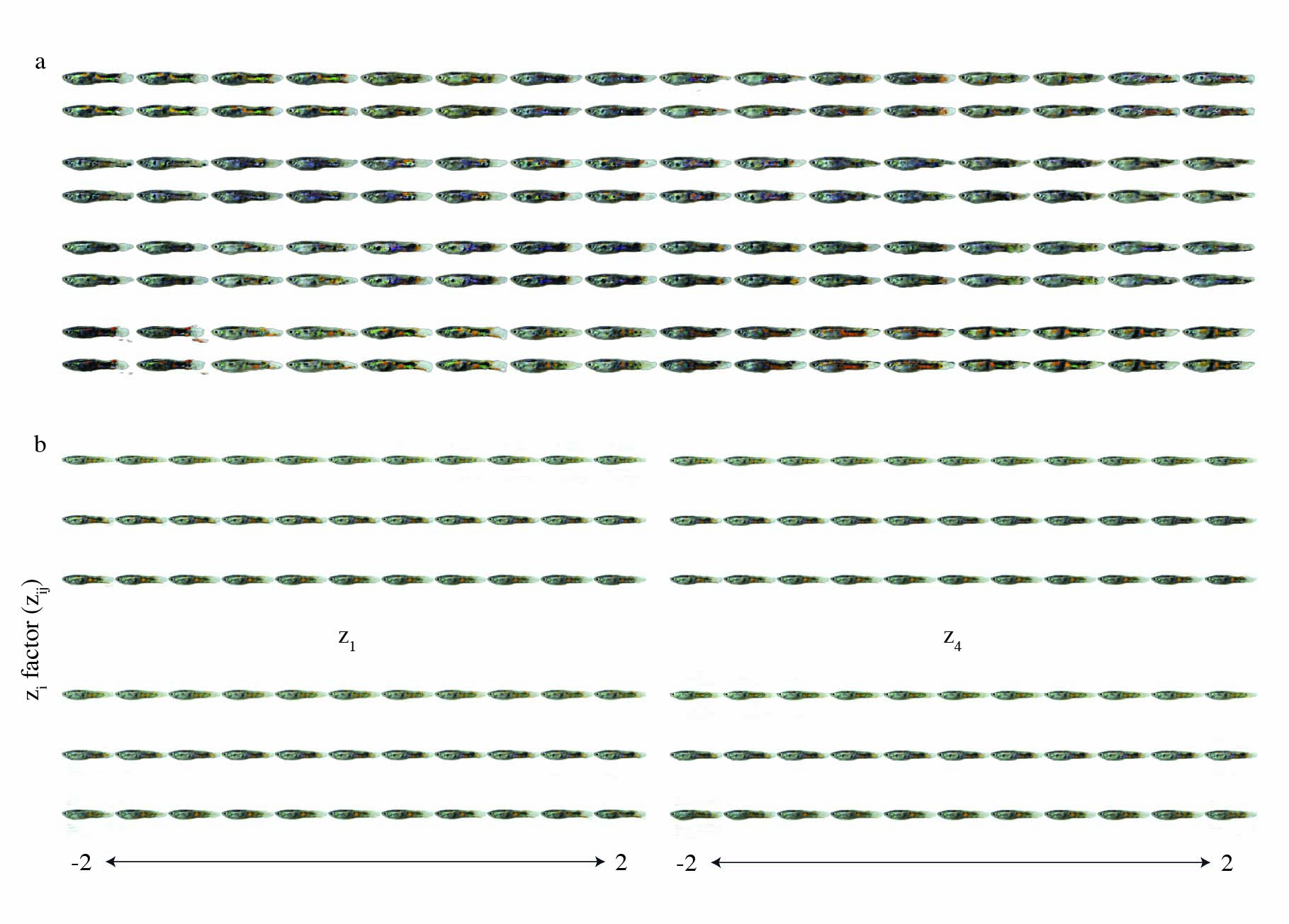}
   \caption{\emph{Exploring latent variables}. a) We incorporated knowledge of our sample data by providing a 32-class categorical latent code and were able to generate examples from distinct classes learned by the model which capture meaningful combinations of features in our sample data. b) Latent traversal of 3 latent variables from $z_1,\dots,z_4$. Top and middle are two embedded samples and bottom a latent code initialized at zero. For each latent variable (rows) we traverse values between -2 and 2 for the generated output. We see that each latent code has consistent effects. All samples in both a and b are generated from the generative models (b and c in Figure \ref{fig:methods}}
   \label{fig:vlae_trav}
\end{figure}

\begin{figure}[ht]
   \includegraphics[width=\textwidth]{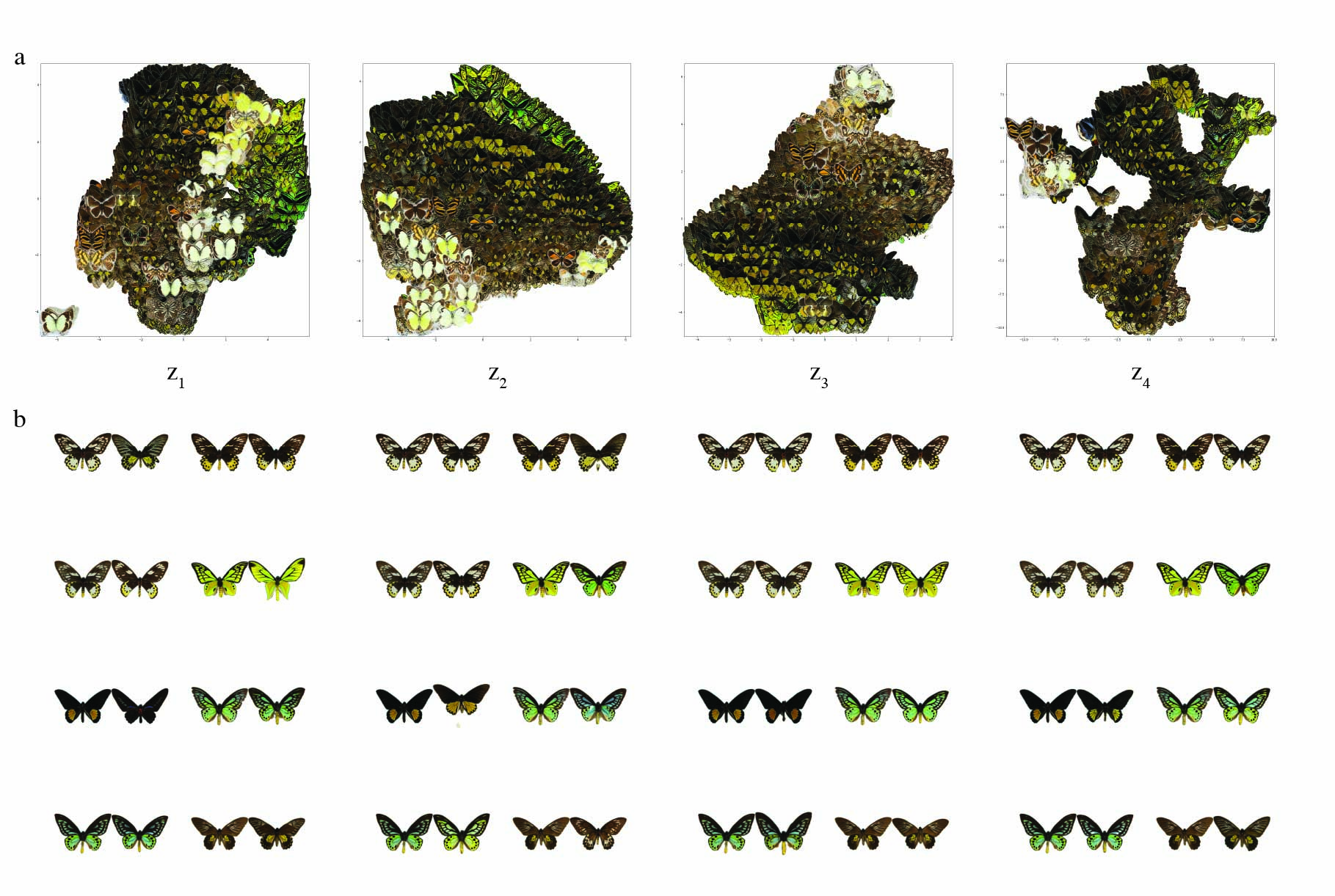}
   \caption{\emph{Exploring latent representations}. a) 2D embedding (using tSNE\cite{vanderMaaten2008, Kobak2019}) of butterfly images using the 4 hierarchical latent encodings. The relationship between images at lower levels are dominated by color value similarities whereas at higher layers pattern elements at increasing spatial scales define the relationship between samples b) Nearest neighbors of 8 random samples based in the Minkowski distance\cite{Dubes1980} between the 10-dimensional space of each latent code $z_1,\dots,z_4$}
   \label{fig:vlae_topo}
\end{figure}

\begin{figure}[ht]
   \centering
   \includegraphics[width=\textwidth]{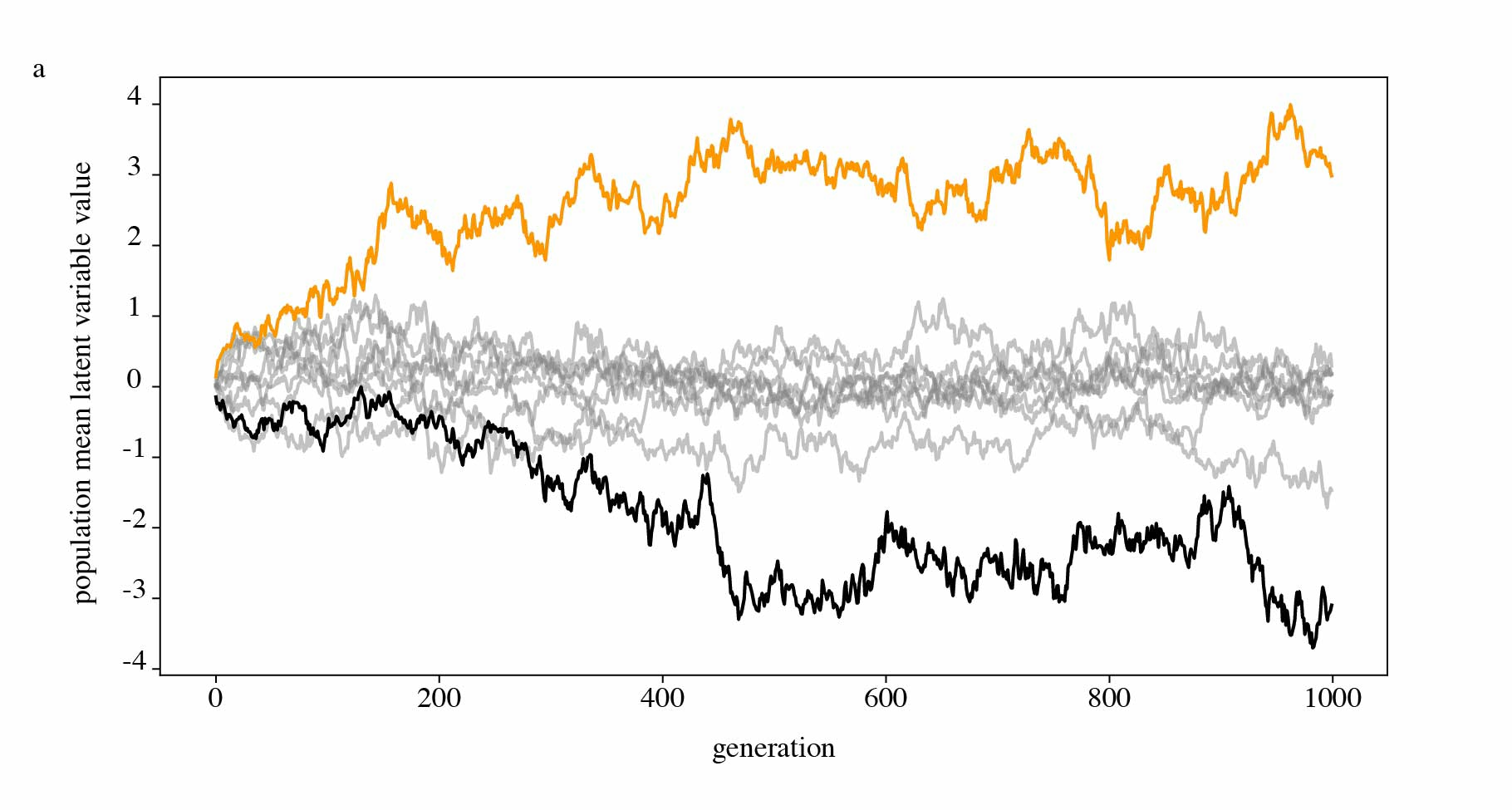}
   \caption{\emph{Latent variable "alleles" over generations} a) We find two alleles are driven to fixation in the population after several generations selecting for oranger and higher contrast color patterns in guppies.}
   \label{fig:alleles}
\end{figure}

\begin{figure}[ht]
   \centering
   \caption{\emph{Movie 1: The combined pattern space over 500 generations, visualized in 2D using tSNE}}
   \label{fig:evospacemovie}
\end{figure}
 
\begin{figure}[ht]
  \centering
  \caption{\emph{Movie 2: VR animation of learned coloration pattern models to an animated guppy for virtual playback.}}
  \label{fig:vrmovie}
\end{figure}

\end{appendices}

\end{document}